\pdfoutput=1

\PassOptionsToPackage{table}{xcolor}  
\documentclass[11pt]{article}

\usepackage[preprint]{acl}

\usepackage{times}
\usepackage{latexsym}
\usepackage{algorithm}
\usepackage{algorithmic}

\usepackage[T1]{fontenc}
\usepackage[utf8]{inputenc}
\usepackage{microtype}
\usepackage{inconsolata}
\usepackage{booktabs}

\usepackage{graphicx}
\usepackage{multirow}

\usepackage{amsmath}
\usepackage{amsthm}
\usepackage{amsfonts}

\definecolor{blue_custom}{RGB}{153, 179, 238} 
\definecolor{mylightblue}{rgb}{0.8, 0.9, 1.0}

\newtheorem{theorem}{Theorem}

\makeatletter
\renewcommand{\maketag@@@}[1]{\hbox{\m@th\normalsize\normalfont#1}}%
\makeatother

\title{Mutual-Taught for Co-adapting Policy and Reward Models}

\author{Tianyuan Shi\textsuperscript{\rm 1}, Canbin Huang\textsuperscript{\rm 1}, Fanqi Wan\textsuperscript{\rm 1}, Longguang Zhong\textsuperscript{\rm 1}, Ziyi Yang\textsuperscript{\rm 1}
\\ \textbf{Weizhou Shen}\textsuperscript{\rm 2}\textbf{,} \textbf{Xiaojun Quan}\textsuperscript{\rm 1}\thanks{$\;\;$Corresponding authors.}\textbf{,} \textbf{Ming Yan\textsuperscript{\rm 2}}
         \\ \textsuperscript{\rm 1}School of Computer Science and Engineering, Sun Yat-sen University, China\\ 
         \textsuperscript{\rm 2}Alibaba Group\\
         \{shity6, huangcb3, wanfq, zhonglg5, yangzy39\}@mail2.sysu.edu.cn \\ quanxj3@mail.sysu.edu.cn, \{shenweizhou.swz, ym119608\}@alibaba-inc.com}
         
\begin{document}
\maketitle
\begin{abstract}
During the preference optimization of large language models (LLMs), distribution shifts may arise between newly generated model samples and the data used to train the reward model (RM). This shift reduces the efficacy of the RM, which in turn negatively impacts the performance of the policy model (PM). To address this challenge, we propose \textbf{Mutual-Taught}, a self-training method that iteratively improves both the PM and RM without requiring additional human annotation. Our approach mirrors the expectation-maximization (EM) algorithm. In the E-step, the PM is updated using feedback from the current RM, guiding the PM toward a better approximation of the latent optimal preference distribution.
In the M-step, we update the RM by constructing training data from the outputs of the PM before and after the E-step update. This process ensures that the RM adapts to the evolving policy distribution. Experimental results demonstrate that this iterative approach leads to consistent improvements in both models. Specifically, our 8B policy model, Llama-3-8B-Instruct-MT, achieves a length-controlled win rate of 54.1\% on AlpacaEval-2, while our 8B reward model, FsfairX-Llama3-RM-MT, performs on par with GPT-4o-2024-08-06 on RewardBench. Our code is available at \url{https://github.com/Stycoo/Mutual-Taught}.

\end{abstract}

\begin{figure}[t]
    \centering
    \includegraphics[width=0.8\linewidth]{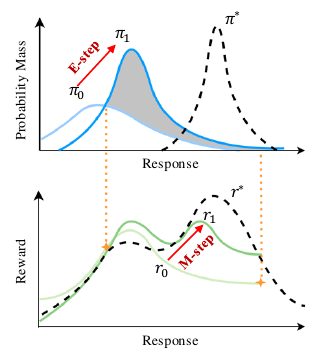}
    \caption{
    An illustration of the Mutual-Taught intuition. The top represents the evolving policy model distribution \(\pi_i\), and the bottom shows the reward model’s preference estimates \(r_i\). After the policy update (E-step), the refined policy model \(\pi_1\) exhibits a higher probability of generating high-reward responses compared to the previous policy \(\pi_0\), as indicated by the shaded region. These improvements are used to enhance the reward model's ability (M-step) to provide more reliable feedback in high-reward regions. Over iterative E-step and M-step, both the policy and reward models progressively adapt and approach their optimal distributions (\(\pi^*\), \(r^*\)).}
    \label{fig:intuition}
    \vspace{-0.4cm}
\end{figure}

\vspace{-0.2cm}
\section{Introduction}

As large language models (LLMs) are fine-tuned to align with human preferences using techniques like reinforcement learning from human feedback (RLHF) \cite{InstructGPT} and Direct Preference Optimization (DPO) \cite{DPO}, the distribution of outputs generated by the evolving policy model may diverge from that of the preference data used to train the reward model. 
This distribution shift leads to a phenomenon known as \emph{reward hacking} \cite{gao2023scaling, zheng2024improving}: as the model adapts, it generates outputs that score well under the current reward model but fail to reflect true human preferences, ultimately compromising alignment reliability.


To address this issue, one potential solution is to continuously gather new human preference annotations for recently generated samples and update the reward model accordingly \cite{Llama2}. However, this approach is not scalable due to its heavy reliance on human labor. An alternative strategy leverages LLM-as-a-Judge prompting \cite{Self-rewarding, Meta-rewarding}, where the LLM evaluates the quality of its own generated outputs and iteratively undergoes DPO training. While this method enhances both the instruction-following and judgment capabilities of the LLM, it requires strong base models or pre-training on judgment-related datasets to develop reliable judgment skills. 

In this paper, we explore methods to mutually improve both the policy and reward models during LLM alignment without relying on external supervision. Our primary research question is: \emph{How can we automatically generate high-quality feedback from LLM alignment to update the reward model, ensuring that its distribution remains consistent with the policy model's distribution?} To address this question, we introduce a self-training framework, termed \textbf{Mutual-Taught}, which is analogous to the expectation-maximization (EM) algorithm, as illustrated in Figure~\ref{fig:intuition}. Specifically, the E-step focuses on optimizing the policy model to achieve better preference alignment with human preferences using the current reward model. In the M-step, the reward model is updated using comparison data derived from the policy's outputs before and after the E-step update. These pseudo-preference pairs naturally emerge from the evolving policy distribution, which eliminates the need for external feedback to update the reward model.

In our experiments, Mutual-Taught leverages Llama-3-8B-Instruct \cite{dubey2024Llama} as the base policy model (PM) and FsfairX-Llama3-RM-v0.1 \cite{xiong2024iterative} as the base reward model (RM). Experimental results demonstrate that iterative training on the UltraFeedback dataset \cite{Cui2024UltraFeedbackBL} leads to substantial improvements in both the PM and RM. For the PM, it achieves a +31.0 LC win rate on AlpacaEval-2 \cite{AlpacaEval} and a +17.8 win rate on Arena-Hard \cite{ArenaHard} over the base model. For the RM, it elevates performance to match GPT-4o-2024-08-06 on RewardBench \cite{RewardBench}.  
Moreover, Mutual-Taught surpasses advanced baselines such as Iterative DPO \cite{dong2024rlhf}, Meta-Rewarding \cite{Meta-rewarding}, and SPPO \cite{SPPO}, emphasizing the critical role of reward model updates during policy optimization.~Overall, these results confirm that mitigating the distributional shift between the reward model and the evolving policy model enhances preference optimization.



\section{Related Work}
\vspace{-0.1cm}
\paragraph{Offline preference optimization}
Reinforcement learning from human feedback (RLHF) \cite{InstructGPT} has emerged as a pivotal approach of preference optimization. However, it depends on reinforcement learning techniques such as proximal policy optimization (PPO) \cite{PPO}, which are challenging to implement and often unstable during training. To address these limitations, Direct Preference Optimization (DPO) \cite{DPO} reparameterizes the reward function in RLHF to directly learn a policy model from preference data, eliminating the need for an explicit reward model and simplifying the training process.
Besides DPO, various preference optimization objectives have been proposed to improve performance and simplify training, including IPO \cite{IPO}, KTO \cite{KTO}, and SimPO \cite{SimPO}. However, without an external reward model, these methods may face challenges in generalization, scalability, and adaptability, increasing the risk of overfitting and misalignment with human preferences.
\paragraph{Iterative preference optimization}  
To enable the policy to learn from data generated by the evolving policy, recent studies have extended preference optimization to an iterative training framework. This approach continuously updates the reference model, either by incorporating the most recent policy model or by generating preference pairs scored and selected by the evolving policy model. For instance, \citeauthor{PCO} \citeyearpar{PCO} propose iterative preference optimization using the Pairwise Cringe Loss (PCO) and generalize DPO to iterative DPO. Analogous to our work, $\text{ReST}^\text{\emph{EM}}$ \cite{ReSTEM} also introduces a self-training method based on expectation-maximization (EM). 
However, $\text{ReST}^\text{\emph{EM}}$ primarily focuses on iteratively optimizing the policy model by generating improved responses for fine-tuning, whereas our method aims to mutually improve both the policy and reward models to address the distribution shift problem.
Other approaches, such as SELM \cite{SELM} and XPO \cite{XPO}, enhance the DPO objective with an optimism-driven exploration term, enabling the model to maintain the ability to explore potentially high-reward policy space during online alignment. SPIN \cite{SPIN}, DNO \cite{DNO}, and SPPO \cite{SPPO} employ a self-play mechanism to refine the policy model using self-generated responses,  bypassing the need for human annotation.

However, these approaches overlook distribution shifts, which can limit the effectiveness of preference alignment. To address distribution shifts, \citeauthor{InstructGPT} \citeyearpar{InstructGPT} collect new responses from the current best policy. These responses are annotated by humans and subsequently used to train a new reward model. While effective, this process incurs significant annotation costs. ReST-MCTS* \cite{ReST-MCTS*} leverages a modified Monte Carlo Tree Search to generate solutions using the policy and evaluates them against ground truth for reward model training. However, its dependence on ground truth restricts its applicability to only a limited set of scenarios.  
In contrast, Self-Rewarding \cite{Self-rewarding} and Meta-Rewarding \cite{Meta-rewarding} adopt an LLM-as-a-Judge mechanism \cite{LLM-as-a-Judge}, where the policy model evaluates its own responses, obviating the need for a separate reward model. However, while this approach simultaneously improves both response generation and evaluation capabilities of the LLM through iterative updates, it relies heavily on strong base models or pretraining on judgment-specific datasets to ensure reliable judgment skills.


\begin{figure*}[htbp]
    \centering
    \includegraphics[width=0.98\linewidth]{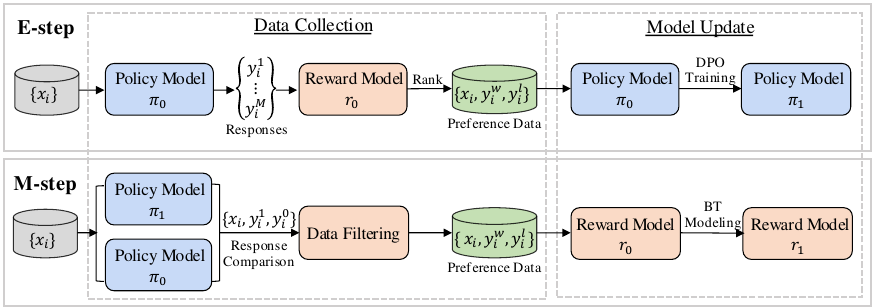}
    \caption{Overview of the Mutual-Taught framework, which alternates between policy model updates (E-step) and reward model updates (M-step). The policy is fine-tuned using reward model feedback (E-step), while the reward model adapts via contrastive comparisons of policy outputs (M-step), requiring no additional human annotations.}
    \label{fig:method}
    \vspace{-0.2cm}
\end{figure*}


\vspace{-0.05cm}
\section{Preliminaries}
\vspace{-0.05cm}
\subsection{Reward Modeling}

In reinforcement learning from human feedback (RLHF) \cite{InstructGPT}, a reward model \( r(y; x) \) is first trained to predict a human preference score for a response \( y \) given a prompt \( x \). This reward model is typically trained using human-annotated preference pairs \((x, y_w, y_l)\), where \( y_w \) is preferred over \( y_l \) for the given prompt \( x \).  
The Bradley-Terry model \cite{BTModel} is widely used to estimate the probability that one response is preferred over another in scenarios where pairwise comparisons are involved:
\begin{equation}
\setlength{\abovedisplayskip}{6pt}
\scalebox{0.91}{$
\begin{aligned}
P(y_w &\succ y_l \mid x) = \sigma(r(y_w; x) - r(y_l; x)) \\
    &= \frac{\exp(r(y_w; x))}{\exp(r(y_w; x)) + \exp(r(y_l; x))}
\end{aligned}
$}
\end{equation}
where $\sigma$ is the sigmoid function. The reward model is trained by maximizing the log-likelihood of observed preferences based on the given equation.


\subsection{Direct Preference Optimization}
Direct Preference Optimization (DPO) \cite{DPO} simplifies the training process by replacing the two-step procedure of RLHF with a single unified objective that directly leverages preference data.
Specifically, DPO derives its objective by reinterpreting preference comparisons with a probabilistic model. This results in a closed-form expression for the optimization objective, where the loss function encourages the model to assign higher probabilities to preferred outputs relative to less-preferred ones, without the need for explicit reward modeling or reinforcement learning:
\begin{equation}
\setlength{\abovedisplayskip}{5pt} 
\scalebox{0.88}{$
\begin{aligned}
&\mathcal{L}_{\text{DPO}}  =  \\ & -\log \sigma \left( \beta \log  \frac{\pi_{\theta}(y_w\!\mid\!x)}{\pi_{\text{ref}}(y_w\!\mid\!x)}  - \beta \log  \frac{\pi_{\theta}(y_l\!\mid\!x)}{\pi_{\text{ref}}(y_l\!\mid\!x)} \right).
\end{aligned}
$}
\end{equation}

However, while DPO offers enhanced stability and ease of optimization by directly leveraging preference data, its offline nature and the absence of an explicit reward model limit its ability to dynamically adapt to changes in the evolving policy distribution. Instead, this work adopts an iterative DPO setup with on-policy sampling and an external reward model for preference annotation.

\section{Mutual-Taught}
\vspace{-0.13cm}
Current on-policy preference optimization methods often assume that the reward model functions as a fixed oracle encoding an ``optimal'' preference distribution. However, this assumption fails in practice as the policy evolves through optimization, causing its output distribution to shift \cite{Llama2, APO}. In such cases, a static reward model trained on outdated data may no longer accurately reflect the optimality. This misalignment results in feedback that increasingly strays from the policy’s true performance.
\vspace{-0.1cm}
\subsection{Overview}
To tackle this challenge, we propose a self-training framework, \textbf{Mutual-Taught}, that co-optimizes both the policy and the reward model. Inspired by the expectation-maximization (EM) algorithm, Mutual-Taught models the latent optimal preference distribution as a hidden variable that evolves over time. The framework iteratively refines both models to approximate and align with this latent distribution in two key phases. \emph{E-Step}: The policy is optimized to better approximate the latent optimal preference distribution, guided by the reward model's current representation of preferences.
\emph{M-Step}: The reward model is updated to reflect the policy's improved outputs, ensuring it remains aligned with the policy's evolving distribution.

As illustrated in Figure~\ref{fig:method}, this co-evolving process enables the policy to progressively generate higher-quality responses while the reward model refines its evaluation criteria accordingly. Consequently, Mutual-Taught can adapt to distributional shifts between the policy and the reward model without requiring additional human annotations.

\subsection{Objective of Mutual-Taught}
Let \( \mathcal{D} \) be a dataset of prompts. For each prompt \( x \in \mathcal{D} \), we assume there exists a latent “optimal” response distribution \( \pi^*(y\!\mid\!x) \), which best reflects true human preferences but is unknown in practice. 
Our objectives are twofold: first, to learn a policy model $\pi_\theta(y\!\mid\!x) $ that approximates the optimal distribution \( \pi^*(y\!\mid\!x) \) through preference learning, guided by a reward model \( r(y; x) \); and second, to optimize the reward model \( r(y; x) \), ensuring that it evaluates responses \( y \) in alignment with \( \pi^*(y\!\mid\!x) \) by leveraging feedback from policy updates. 
We frame this as maximizing the expected reward under the latent optimal distribution:
\begin{equation}
\max_{\pi, r} \mathbb{E}_{x \sim \mathcal{D}, y \sim \pi^*(\cdot \mid x)}[r(y; x)].
\end{equation}
Since \( \pi^*(y\!\mid\!x) \) is unknown, we regard it as a latent distribution and approximate it by updating both the policy and the reward model. In the EM framework, this involves alternating between updating $\pi_\theta(y\!\mid\!x)$ (E-step) and $r(y; x)$ (M-step) to progressively align the policy with \( \pi^*(y\!\mid\!x) \). 

\textbf{E-step}:~This step can be implemented using various preference optimization methods such as RLHF \cite{InstructGPT} and DPO \cite{DPO}. In this work, we illustrate this using DPO for its simplicity and effectiveness.
Assuming the reward model in iteration \( t \) is \( r_{t-1} \), the E-step updates the policy \( \pi_{t-1} \) to \( \pi_t \) by solving:
\begin{equation}
\label{eq:e-step}
\scalebox{0.86}{$
\begin{aligned}
&\pi_t  =  \operatorname*{argmax}_{\pi} \, \mathbb{E}_{x \sim \mathcal{D}_{t}} \\
    & \bigg[\log \sigma \bigg( \beta \log \frac{\pi_\theta(y_w\!\mid\!x)}{\pi_{t-1}(y_w\!\mid\!x)} - \beta \log \frac{\pi_\theta(y_l\!\mid\!x)}{\pi_{t-1}(y_l\!\mid\!x)} \bigg) \bigg],
\end{aligned}
$}
\end{equation}
where $\pi_{t-1}$ acts as the reference model, \( y_w \) and \( y_l \) represent chosen and rejected responses, respectively, both sampled from \(\pi_{t-1}\) and ranked by \( r_{t-1}\). 

\textbf{M-step}:~After obtaining \(\pi_t\), we fix it and update the reward model \( r_{t-1}\) to \( r_t\). For a given prompt \( x \), let \( y_{t-1} \) and \( y_t \) be the responses generated by \( \pi_{t-1} \) and \( \pi_t \), respectively. Since \( \pi_t \) is optimized with respect to \( r_{t-1} \), we treat \( y_t \) as the preferred response relative to \( y_{t-1} \). We then construct pseudo-preference pairs \( (y_t, y_{t-1}) \) and update \( r_{t-1}\) by maximizing the Bradley-Terry log-likelihood:
\begin{equation}
\label{eq:m-step}
\scalebox{0.95}{$
\displaystyle
r_{t} = \operatorname*{argmax}_{r} \; \mathbb{E}_{x \sim \mathcal{D}_{\text{R}}} \left[ \log P_r(y_t \succ y_{t-1} \mid x) \right]
$}
\end{equation}
The M-step ensures the reward model remains accurate in distinguishing responses generated by \( \pi_t \).
\vspace{-0.05cm}
\subsection{Two-Stage Stabilization}
\label{subsec:two_stage}
\vspace{-0.05cm}
While the EM framework provides theoretical convergence guarantees under certain conditions (see Appendix \ref{appendix:theoretical_convergence}), practical implementations face two challenges in the iterative learning process: (1) \textit{Policy degradation risk} due to over-optimization in the E-step, and (2) \textit{Reward distortion} arising from noisy pseudo-labels in the M-step. To address these challenges, we propose a two-stage stabilization.
\vspace{-0.1cm}
\paragraph{Model selection for E-step}\hspace{-0.1cm}To prevent potential policy degradation in the E-step, we implement a validation-based model selection strategy. Specifically, we evaluate the policy checkpoints \( \{\pi^{k}_t\} \) saved in the $t$-th iteration against \( \pi_{t-1} \) from the previous iteration on a fixed validation set \( \mathcal{D}_\text{MS} \). The win rate for each checkpoint is computed as:
\begin{equation}
\scalebox{1.1}{$
w^{k}_t = \frac{1}{|\mathcal{D}_\text{MS}|} \sum_{x \in \mathcal{D}_\text{MS}} \mathbb{I}\left(y^{k}_t \succ y_{t-1} \mid x\right)
$}
\end{equation}
where $y_{t-1} \sim \pi_{t-1}(\cdot|x)$, $y^{k}_t \sim \pi^k_t(\cdot|x)$, and $\mathbb{I}(\cdot)$ is an indicator function defined as:
\begin{equation}
\label{eq:indicator}
\scalebox{0.85}{$
\mathbb{I}\left(y^{k}_t \succ y_{t-1} \mid x\right) =
\begin{cases}
1 & \text{if } r_{t-1}(y^{k}_t; x) > r_{t-1}(y_{t-1}; x), \\
0 & \text{otherwise}.
\end{cases} \nonumber
$}
\end{equation}
Only the checkpoint that demonstrates maximum improvement over the previous policy is selected, thereby ensuring monotonic policy enhancement:
\begin{equation}
\label{eq:model_sel}
    \pi_t = \operatorname*{argmax}_{\pi^k_t} w^k_t.
\end{equation}
If no candidate in this iteration demonstrates sufficient improvement (\( \max_k w^{k}_t < \tau \)), the iteration halts, and the previous model is preserved.
\vspace{-0.1cm}
\paragraph{Data filtering for M-step} 
To mitigate the impact of unreliable preference pairs that could distort reward learning, we implement dynamic data filtering in the M-step to remove noisy pseudo-labels \cite{huang2022UPL}. We first compute the reward margin for each pseudo-pair \( (y_t, y_{t-1}) \) as follows:
\begin{equation}
\label{eq:data_filtering}
\Delta r(x) = r_{t-1}(y_t \, ; \, x) - r_{t-1}(y_{t-1} \, ; \, x)
\end{equation}

To adaptively filter noisy comparisons, we establish a variance-aware threshold $\epsilon_t = \sqrt{\mathbb{V}_{x\sim\mathcal{D}_R}[r_{t-1}(y_{t-1};x)]}$ that automatically adjusts to the reward model's uncertainty level \cite{westofn}. 
Only pairs satisfying $|\Delta r(x)| \geq \epsilon_t$ are considered high-confidence pseudo-labels. Our filtering strategy removes pairs with $\Delta r(x) \leq -\epsilon_t$, as they are confidently identified as noisy samples.

Particularly, when \( \Delta r(x) > \epsilon_t \), this strategy selects high-confidence and high-quality samples, which reinforce the RM’s capabilities through self-training. When \( -\epsilon_t < \Delta r(x) < 0 \), these slightly noisy pairs serve as regularization that prevents the RM from overfitting to the policy's distribution. Experimental results show that this data filtering strategy improves both the RM and the policy model. For more details, see Appendix \ref{appendix:different_filter_methods}. 

\section{Experiments}
\vspace{-0.1cm}
\subsection{Experimental Setup}
\paragraph{Base models and training dataset}
We use Llama3-8B-Instruct \cite{dubey2024Llama} as our base policy model and FsfairX-Llama3-RM-v0.1 \cite{xiong2024iterative} as the initial reward model. FsfairX-Llama3-RM is one of the top-performing 8B models on RewardBench \cite{RewardBench} and offers open-source code that facilitates continuous training. Following previous work, we use the UltraFeedback dataset \cite{Cui2024UltraFeedbackBL} for training, which comprises approximately 60,000 prompts from diverse sources. We partition the dataset into three subsets: one for initial policy training, one for reward model updates, and one for policy re-updates. Thus, there are two policy iterations and one reward model iteration in a full round of the dataset. In our practical implementation, we utilize the mixed preference data from the first and third partitions to train the reward model. Refer to Section \ref{sec:fur_analysis} and Appendix \ref{Experiments Details} for more details. 

\paragraph{Evaluation benchmarks}
In order to investigate whether the policy model and the reward model can mutually enhance each other through our Mutual-Taught, we conduct separate evaluations of each model. For \emph{policy evaluation}, we utilize two widely recognized automatic evaluation benchmarks, AlpacaEval-2 \cite{AlpacaEval} and Arena-Hard \cite{ArenaHard}, with GPT-4\footnote{In AlpacaEval-2, GPT-4-Preview-1106 serves as both the baseline and the judge. In Arena-Hard, GPT-4-0314 serves as the baseline, while GPT-4-Preview-1106 acts as the judge.} serving as the judge. Each benchmark targets different aspects of model performance. AlpacaEval-2 assesses chat capabilities using 805 instructions spanning a wide range of prompts, evaluated through length-controlled (LC) win rate and raw win rate (WR) metrics. Arena-Hard presents more challenging tasks, including 500 well-defined technical problem-solving queries.~For \emph{reward model evaluation}, we assess the reward model's accuracy using RewardBench~\cite{RewardBench}, which measures performance across four categories: Chat, Chat-Hard, Safety, and Reasoning.
\vspace{-0.1cm}
\paragraph{Baselines}

We evaluate our method against a variety of baselines, including \textit{offline preference optimization} and \textit{iterative preference optimization} methods. Refer to Appendix \ref{Baselines} for more details. 


\begin{table*}[htbp]
\centering
\resizebox{0.85\textwidth}{!}{
\begin{tabular}{lcccccc}
\hline
\multicolumn{1}{l}{\multirow{2}{*}{\textbf{Model}}} & \multicolumn{3}{c}{\textbf{AlpacaEval-2}} & \multicolumn{2}{c}{\textbf{Arena-Hard}} \\
\multicolumn{1}{c}{}  & \textbf{LC Win Rate}  & \textbf{Win Rate}  & \textbf{Avg. Len}  & \textbf{Win Rate} & \textbf{Avg. Len} \\ 
\hline
\multicolumn{6}{c}{\textit{Base Policy Model}} \\
\hline
Llama-3-8B-Instruct     & 23.1        & 23.1     & 1899    & 20.6     & 585 \\ 
\hline
\multicolumn{6}{c}{\textit{Offline Preference Optimization Methods}} \\
\hline
SimPO                 & 47.9        & 46.3     & 1934   & 32.5     & 552 \\ 
IPO                   & 43.7        & 42.1     & 1899   &  34.5    & 569  \\
 DPO                   & 44.3        & 42.7     & 1945   & 33.1     & 557 \\ \hline
\multicolumn{6}{c}{\textit{Iterative Preference Optimization Methods}} \\
\hline
Meta-Rewarding Iter1  & 34.2        & 32.6     & 1893    & 27.7     & 531 \\
Meta-Rewarding Iter2  & 36.4        & 34.5     & 1876    & 27.0     & 530 \\
Meta-Rewarding Iter3  & 37.5 $\textcolor{blue}{(\uparrow 14.4)}$       & 35.2 $\textcolor{blue}{(\uparrow 12.1)}$  & 1868     & 27.9 $\textcolor{blue}{(\uparrow 7.3)}$  & 530 \\ 
\hline
SPPO Iter1            & 39.4        & 39.5     & 2021    & 30.6     & 570 \\
SPPO Iter2            & 41.0        & 44.4     & 2396    & 34.4     & 653 \\
SPPO Iter3            & 46.4 $\textcolor{blue}{(\uparrow 23.3)}$       & 48.5  $\textcolor{blue}{(\uparrow 25.4)}$   & 2128  & 33.6 $\textcolor{blue}{(\uparrow 13.0)}$    & 542 \\ 
\hline
DPO Iter1            & 33.6        & 33.8     & 1989    & 30.3     & 559 \\
DPO Iter2            & 43.4        & 42.3     & 1961    &  33.3    & 587 \\
DPO Iter3            & 47.2 $\textcolor{blue}{(\uparrow 24.1)}$     & 48.7  $\textcolor{blue}{(\uparrow 25.6)}$   & 1930   &  34.7 $\textcolor{blue}{(\uparrow 14.1)}$   & 571 \\ 
\hline
\multicolumn{6}{c}{\textit{Our Methods}} \\
\hline
\rowcolor{mylightblue} Mutual-Taught Iter1   & 38.4        & 37.3     & 1943    & 33.9     & 549 \\
\rowcolor{mylightblue} Mutual-Taught Iter2   & \textbf{54.1} $\textcolor{blue}{(\uparrow 31.0)}$ & \textbf{55.9} $\textcolor{blue}{(\uparrow 32.8)}$    & 2177    & \textbf{38.4} $\textcolor{blue}{(\uparrow 17.8)}$  & 682 \\ 
\hline
\end{tabular}
}
\caption{Overall results of our proposed Mutual-Taught method with Llama-3-8B-Instruct as the policy model, compared against various baseline methods on AlpacaEval-2 and Arena-Hard. Text in \textbf{bold} indicates the best performance. The numbers in brackets represent the degree of improvement relative to Llama-3-8B-Instruct. }
\vspace{-0.2cm}
\label{tab:main-results}
\end{table*}
\vspace{-0.1cm}
\subsection{Main Results}
\paragraph{Iterative performance improvement on policy} In Table \ref{tab:main-results}, we report the performance of Mutual-Taught and baseline methods on the instruction-following benchmarks, AlpacaEval-2 and Arena-Hard. Mutual-Taught shows substantial improvements to the Llama-3-8B-Instruct model, achieving a 31.0-point increase in length-controlled (LC) win rate on AlpacaEval-2 and a 17.8-point increase in win rate on Arena-Hard, respectively. 
Compared to baseline methods, our method demonstrates clear superiority on both AlpacaEval-2 and Arena-Hard.


\begin{figure}[t]
    \centering
    \includegraphics[width=\linewidth]{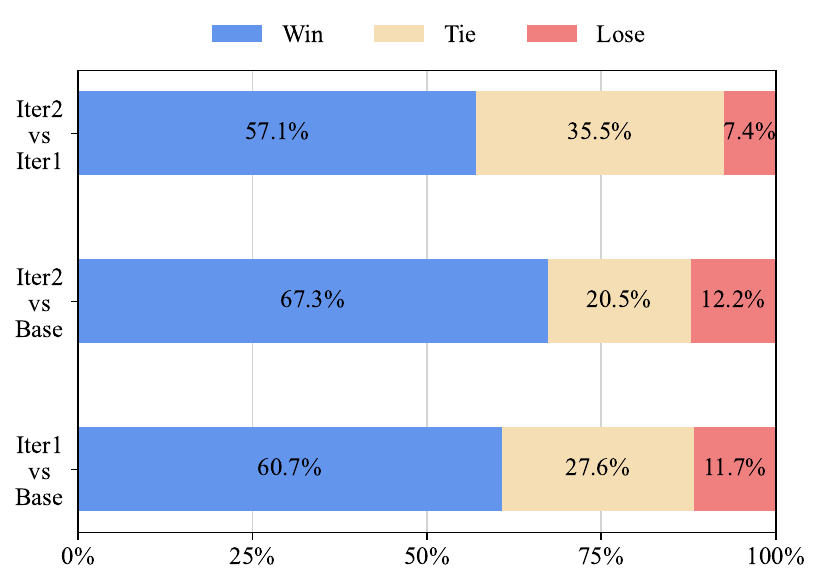}
    \caption{Results of in-distribution (ID) evaluation of reward models obtained through Mutual-Taught. We compare reward models from different iterations, presenting the pairwise win, tie, and lose rates.}
    \vspace{-0.5cm}
    \label{fig:reward-model}
\end{figure}

Note that our method employs only two-thirds of the available datasets for updating the policy model, reserving the remaining for updating the reward model. Despite using less data for policy model iterations compared to other iterative baselines, we achieve notably better performance on AlpacaEval-2 and Arena-Hard. This result highlights the importance of iteratively updating both the policy and reward models during the training process. Moreover, it also suggests that improving the reward model offers greater benefits than just increasing training data for the policy model.

\paragraph{Iterative performance improvement on reward model} To evaluate the effectiveness of Mutual-Taught in enhancing the reward model (RM), we analyze its performance across two scenarios.

\textit{In-distribution} (ID):  
We first assess the RM's performance under ID conditions. Specifically, we use the policy model after two iterations to generate responses for 2000 randomly sampled prompts from the Ultrafeedback test set. The base RM and iteratively updated RMs (from Mutual-Taught) are then tasked with selecting the optimal response, with GPT-4-Preview-1106 serving as the judge for pairwise comparisons. As shown in Figure~\ref{fig:reward-model}, the iteratively updated RMs achieve progressively higher win rates against the base RM, demonstrating their improved ability to identify high-quality responses. This enhancement ensures more reliable training data for subsequent policy iterations.

\begin{table*}[t]
\centering
\resizebox{0.75\textwidth}{!}{
\begin{tabular}{lccccc}
\hline
\textbf{Model}             & \textbf{Chat}  & \textbf{Chat Hard} & \textbf{Safety} & \textbf{Reasoning} & \textbf{Average} \\ \hline
GPT-4o-2024-08-06 & 96.1 & 76.1     & 88.1  & 86.6      & 86.7    \\ 
FsfairX-Llama3-RM-v0.1 & 99.4 & 65.1     & 87.8  & 86.4  & 84.7 \\ \hline 
Mutual-Taught Iter1      & 98.3 & 63.9     & 85.1  & 95.8     & 85.8    \\
Mutual-Taught Iter2      & 98.2 & 66.3     & 87.8  & 95.7      & 87.0    \\
\hline
\end{tabular}}
\caption{Out-of-distribution (OOD) evaluation results of reward models on RewardBench.}
\vspace{-0.25cm}
\label{tab:ood}
\end{table*}

\textit{Out-of-distribution} (OOD): We further evaluate the RM's generalization capability using RewardBench. As shown in Table~\ref{tab:ood}, the RM exhibits consistent improvement after each iteration, with an average score increase of 2.3 points after two iterations, approaching the performance of GPT-4o-2024-08-06. Notably, in the reasoning dimension, the RM achieves a clear performance boost after the first iteration, ultimately attaining a 9.3-point improvement. In other dimensions, the RM initially declines but recovers and stabilizes at the base RM level. This behavior is attributed to the varying initial performance of the policy model (PM) across dimensions, which influences the quality of training data generated by comparing the PM’s outputs before and after each iteration. Specifically, in the reasoning dimension, where the PM has stronger initial performance, the RM receives higher-quality training data, leading to substantial improvements. In other dimensions, the PM's weaker initial performance results in lower-quality training data, causing a temporary decline in RM performance. However, as the PM evolves through iterations, the RM benefits from better-quality data and ultimately leads to improved performance.

\vspace{0.05cm}

\subsection{Further Analysis}
\label{sec:fur_analysis}
\paragraph{Impact of reward model training data type}
\label{Impact of Reward Model Training Data Type}


Our data construction strategy is designed to meet two critical requirements for effective iterative alignment: (1) enabling the reward model to track policy model distribution shifts across iterations, and (2) maintaining stable learning signals throughout policy optimization.~While previous work \cite{westofn} shows that on-policy sampling data annotated by the reward model can enhance its robustness through iterative self-supervision, we argue that explicitly capturing policy evolution via our comparison strategy offers crucial dynamic alignment signals for updating the reward model. To explore this effect, we conduct experiments using three distinct data types to train the reward model: \emph{self-training}, \emph{policy-comparison}, and \emph{mixed}.

The self-training data comprises preference data used in the first iteration of policy model optimization, with labels derived from the base reward model. This preference data reflects the initial capabilities of the reward model. The policy-comparison data is constructed from responses generated by the policy both before and after iteration, capturing shifts in the policy distribution. The mixed data type, which combines both self-training and policy-comparison preference data, aims to leverage the unique strengths of each approach.
\begin{figure}[t]
    \centering
    \includegraphics[width=\linewidth]{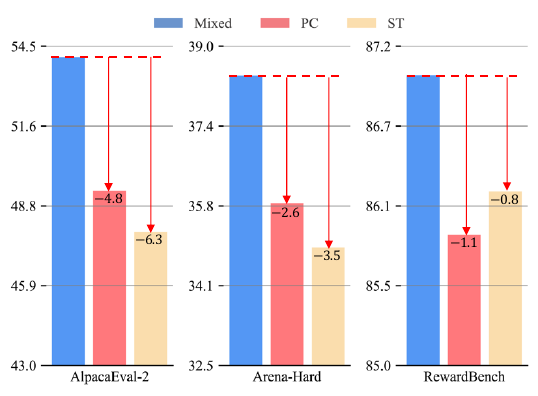}
    \caption{Impact of different reward model training data types on the performance of Mutual-Taught. For brevity, policy-comparison data and self-training data are abbreviated as PC and ST, respectively.}
    \vspace{-0.5cm}
    \label{fig:rm-data-type}
\end{figure}

As shown in Figure \ref{fig:rm-data-type}, the policy model’s performance declines when using either self-training or policy-comparison data in isolation, compared to the mixed preference data. Specifically, when only self-training data is used, the policy model’s performance drops by 6.3 and 3.5 points, respectively, on AlpacaEval and ArenaHard, while the reward model's performance shows no significant decline. In contrast, when only policy-comparison data is used, the reward model performance slightly deteriorates, but the policy model’s performance is less affected. We hypothesize that self-training data, which reflects the reward model's initial distribution, helps prevent catastrophic forgetting but is less effective at capturing improved preference distributions. This limits its ability to guide the policy model in subsequent iterations. On the other hand, policy-comparison data, which compares the updated and previous policy models, aligns more closely with the iterative optimization goal, enabling the reward model to better approximate the improved preference distribution and offer more effective feedback for policy updates. The integration of both data types in Mutual-Taught strikes a balance between preventing knowledge forgetting and modeling improved preference distributions. As a result, Mutual-Taught achieves superior performance compared to using either data type alone.
\paragraph{Performance of Mutual-Taught with additional rounds}
To investigate the effect of extending Mutual-Taught beyond the main experimental setup, we conduct an additional round of training using the same dataset and hyperparameters. Each round consists of two policy model updates and one reward model update. Crucially, to mitigate overfitting from repeated training on the same data, the policy and reward models from the previous round are not directly fine-tuned further. Instead, they are used solely to generate higher-quality training data for the next iteration, with the new iteration’s models starting from the base models. 
\begin{figure}[t]
    \centering
    \includegraphics[width=\linewidth]{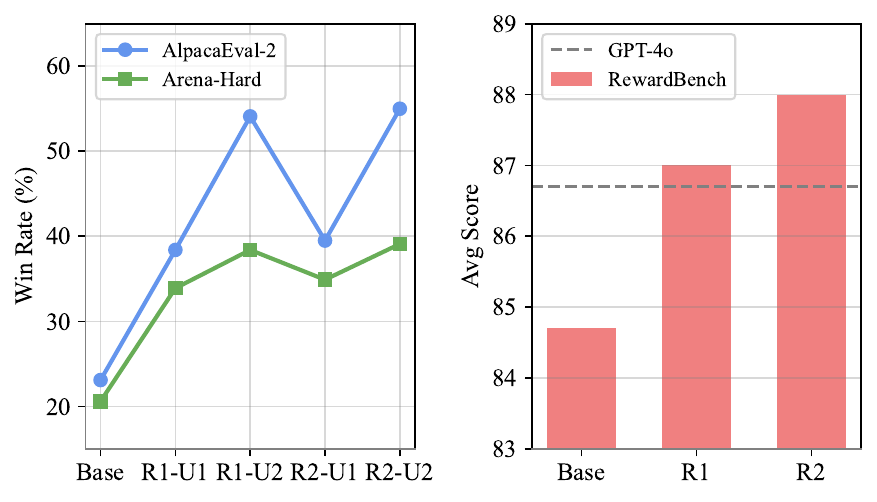}
    \caption{Performance of the policy (left) and the reward (right) models across two rounds. Each round includes two policy updates and one reward model update. For brevity, each policy update is abbreviated (e.g., the first update in Round 1 is denoted as R1–U1).
    }
    \vspace{-0.2cm}
    \label{fig:additional-iterations}
\end{figure}

As shown in Figure \ref{fig:additional-iterations}, both the policy and reward models continue to improve in the second round relative to the first. Notably, the final reward model outperforms GPT-4o-2024-08-06 on RewardBench, demonstrating that Mutual-Taught achieves even better performance with an additional round.
More specifically, in the second iteration, both the policy and reward models utilize preference data generated by their respective fine-tuned predecessors. These higher-quality outputs strengthen the foundation for the E-step (policy updates) and M-step (reward model updates) and result in better alignment between the policy and reward models and enhanced results.

\paragraph{Generalization of the iterated reward model}
In our experiments, the improvement of the reward model depends on training data provided by the policy model (Llama-3-8B-Instruct). Although the final iterated reward model shows performance gains in both in-distribution (ID) and out-of-distribution (OOD) scenarios, it remains unclear whether these improvements can generalize effectively to optimize other policy models. To investigate this, we apply the reward models obtained through the Mutual-Taught iterative process, as reported in the main experiment, to train a different policy model, Mistral-7B-Instruct-v0.2 \cite{jiang2023mistral7b}, using a single iteration of DPO on  UltraFeedback.

\begin{table}[t]
\centering
\resizebox{\linewidth}{!}{
\begin{tabular}{lcc}
\hline
\multirow{2}{*}{\textbf{Model}}  & \multicolumn{2}{c}{\textbf{AlpacaEval-2}} \\
                        & \textbf{LC Win Rate} & \textbf{Win Rate} \\
\hline
Mistral-7B-Instruct-v0.2   & 19.4    & 15.8 \\
~~~~w/ RM-Base          & 42.0       & 42.8  \\
~~~~w/ RM-Iter1         & 45.5       & 45.0   \\
~~~~w/ RM-Iter2         & \textbf{46.8}       & \textbf{51.0}  \\
\hline
\end{tabular}
}
\caption{Effect of the generalization of reward models obtained from Mutual-Taught’s iterative process on guiding the DPO training of Mistral-7B-Instruct-v0.2.}
\vspace{-0.2cm}
\label{tab:generalization}
\end{table}

As shown in Table~\ref{tab:generalization}, using the iterated reward models boosts the policy model's performance on AlpacaEval-2 by up to 4.8 points compared to the base reward model. This demonstrates that the improved reward models, fine-tuned by a specific policy model during the Mutual-Taught iterative process, are not limited to that policy model but can generalize to others. 
The effectiveness of this generalization stems from the fact that the iterated reward models, fine-tuned with improved preference data generated by the evolving policy model, learn a more robust understanding of what constitutes an optimal response. This enhanced capability allows them to provide valuable feedback not only for the policy model they were originally trained with but also for other models on the same task.

\section{Conclusion}

This paper introduces Mutual-Taught, a novel co-evolving framework designed to address the distributional shift challenge in preference learning. Mutual-Taught enables the collaborative improvement of both policy and reward models through an expectation-maximization (EM)-inspired approach, with a dynamic feedback loop between policy optimization (E-step) and reward calibration (M-step). Empirical results show that this iterative process consistently enhances both the policy and reward models. The resulting policy model outperforms existing methods, such as DPO, SPPO, and Meta-Rewarding, across multiple benchmarks, including AlpacaEval-2 and Arena-Hard. Furthermore, the iterated reward model performs on par with GPT-4o-2024-08-06 on RewardBench. These findings confirm that addressing the distributional shift between the reward model and the evolving policy model facilitates further preference optimization.


\section*{Limitations}
Mutual-Taught relies on iterative optimization and feedback during the training of a policy model. However, when applied to tasks involving complex logical reasoning and long-term dependencies, it may face challenges such as slow convergence. Moreover, over-optimization may occur if iterations are allowed to continue without limit.

\section*{Ethics Statement}
All the experiments in this study were conducted using publicly available datasets that do not contain any private or offensive information. Our work does not involve the analysis or utilization of identity characteristics, nor does it engage in any form of gender, racial, or other discrimination.

\section*{Acknowledgements}
This work was supported by the National Natural Science Foundation of China (No. 62176270) and the Guangdong Basic and Applied Basic Research Foundation (No. 2023A1515012832).

\bibliography{acl}

\begin{thebibliography}{38}
\providecommand{\natexlab}[1]{#1}

\bibitem[{Azar et~al.(2024)Azar, Guo, Piot, Munos, Rowland, Valko, and Calandriello}]{IPO}
Mohammad~Gheshlaghi Azar, Zhaohan~Daniel Guo, Bilal Piot, Remi Munos, Mark Rowland, Michal Valko, and Daniele Calandriello. 2024.
\newblock A general theoretical paradigm to understand learning from human preferences.
\newblock In \emph{International Conference on Artificial Intelligence and Statistics}.

\bibitem[{Beeching et~al.(2023)Beeching, Fourrier, Habib, Han, Lambert, Rajani, Sanseviero, Tunstall, and Wolf}]{open-llm-leaderboard}
Edward Beeching, Clémentine Fourrier, Nathan Habib, Sheon Han, Nathan Lambert, Nazneen Rajani, Omar Sanseviero, Lewis Tunstall, and Thomas Wolf. 2023.
\newblock Open {LLM} leaderboard.

\bibitem[{Bradley and Terry(1952)}]{BTModel}
Ralph~Allan Bradley and Milton~E. Terry. 1952.
\newblock Rank analysis of incomplete block designs: I. the method of paired comparisons.
\newblock \emph{Biometrika}, 39:324.

\bibitem[{Chen et~al.(2024)Chen, Deng, Yuan, Ji, and Gu}]{SPIN}
Zixiang Chen, Yihe Deng, Huizhuo Yuan, Kaixuan Ji, and Quanquan Gu. 2024.
\newblock Self-play fine-tuning converts weak language models to strong language models.
\newblock In \emph{International Conference on Machine Learning}.

\bibitem[{Cheng et~al.(2024)Cheng, Yang, Li, Dai, Hu, Cao, Du, and Li}]{APO}
Pengyu Cheng, Yifan Yang, Jian Li, Yong Dai, Tianhao Hu, Peixin Cao, Nan Du, and Xiaolong Li. 2024.
\newblock Adversarial preference optimization: Enhancing your alignment via rm-llm game.
\newblock In \emph{Findings of the Association for Computational Linguistics ACL 2024}, pages 3705--3716.

\bibitem[{Cobbe et~al.(2021)Cobbe, Kosaraju, Bavarian, Chen, Jun, Kaiser, Plappert, Tworek, Hilton, Nakano, Hesse, and Schulman}]{cobbe2021gsm8k}
Karl Cobbe, Vineet Kosaraju, Mohammad Bavarian, Mark Chen, Heewoo Jun, Lukasz Kaiser, Matthias Plappert, Jerry Tworek, Jacob Hilton, Reiichiro Nakano, Christopher Hesse, and John Schulman. 2021.
\newblock Training verifiers to solve math word problems.
\newblock \emph{arXiv preprint arXiv:2110.14168}.

\bibitem[{Cui et~al.(2024)Cui, Yuan, Ding, Yao, Zhu, Ni, Xie, Liu, and Sun}]{Cui2024UltraFeedbackBL}
Ganqu Cui, Lifan Yuan, Ning Ding, Guanming Yao, Wei Zhu, Yuan Ni, Guotong Xie, Zhiyuan Liu, and Maosong Sun. 2024.
\newblock {UltraFeedback}: Boosting language models with high-quality feedback.
\newblock In \emph{International Conference on Machine Learning}.

\bibitem[{Dong et~al.(2024)Dong, Xiong, Pang, Wang, Zhao, Zhou, Jiang, Sahoo, Xiong, and Zhang}]{dong2024rlhf}
Hanze Dong, Wei Xiong, Bo~Pang, Haoxiang Wang, Han Zhao, Yingbo Zhou, Nan Jiang, Doyen Sahoo, Caiming Xiong, and Tong Zhang. 2024.
\newblock {RLHF} workflow: From reward modeling to online {RLHF}.
\newblock \emph{Transactions on Machine Learning Research}.

\bibitem[{Dubey et~al.(2024)Dubey, Jauhri, Pandey, Kadian, Al-Dahle, Letman, Mathur, Schelten, Yang, Fan et~al.}]{dubey2024Llama}
Abhimanyu Dubey, Abhinav Jauhri, Abhinav Pandey, Abhishek Kadian, Ahmad Al-Dahle, Aiesha Letman, Akhil Mathur, Alan Schelten, Amy Yang, Angela Fan, et~al. 2024.
\newblock The llama 3 herd of models.
\newblock \emph{arXiv preprint arXiv:2407.21783}.

\bibitem[{Ethayarajh et~al.(2024)Ethayarajh, Xu, Muennighoff, Jurafsky, and Kiela}]{KTO}
Kawin Ethayarajh, Winnie Xu, Niklas Muennighoff, Dan Jurafsky, and Douwe Kiela. 2024.
\newblock {KTO}: Model alignment as prospect theoretic optimization.
\newblock In \emph{International Conference on Machine Learning}.

\bibitem[{Gao et~al.(2023)Gao, Schulman, and Hilton}]{gao2023scaling}
Leo Gao, John Schulman, and Jacob Hilton. 2023.
\newblock Scaling laws for reward model overoptimization.
\newblock In \emph{International Conference on Machine Learning}.

\bibitem[{Hendrycks et~al.(2021)Hendrycks, Burns, Basart, Zou, Mazeika, Song, and Steinhardt}]{hendrycks2021measuring}
Dan Hendrycks, Collin Burns, Steven Basart, Andy Zou, Mantas Mazeika, Dawn Song, and Jacob Steinhardt. 2021.
\newblock Measuring massive multitask language understanding.
\newblock In \emph{International Conference on Learning Representations}.

\bibitem[{Huang et~al.(2022)Huang, Chu, and Wei}]{huang2022UPL}
Tony Huang, Jack Chu, and Fangyun Wei. 2022.
\newblock Unsupervised prompt learning for vision-language models.
\newblock \emph{arXiv preprint arXiv:2204.03649}.

\bibitem[{Jiang et~al.(2023)Jiang, Sablayrolles, Mensch, Bamford, Chaplot, de~las Casas, Bressand, Lengyel, Lample, Saulnier et~al.}]{jiang2023mistral7b}
AQ~Jiang, A~Sablayrolles, A~Mensch, C~Bamford, DS~Chaplot, D~de~las Casas, F~Bressand, G~Lengyel, G~Lample, L~Saulnier, et~al. 2023.
\newblock Mistral 7b.
\newblock \emph{arXiv preprint arXiv:2310.06825}.

\bibitem[{K{\"o}pf et~al.(2023)K{\"o}pf, Kilcher, von R{\"u}tte Sotiris Anagnostidis Zhi Rui~Tam et~al.}]{OpenAssistant}
Andreas K{\"o}pf, Yannic Kilcher, Dimitri von R{\"u}tte Sotiris Anagnostidis Zhi Rui~Tam, et~al. 2023.
\newblock Openassistant conversations - democratizing large language model alignment.
\newblock In \emph{Thirty-seventh Conference on Neural Information Processing Systems Datasets and Benchmarks Track}.

\bibitem[{Lambert et~al.(2024)Lambert, Pyatkin, Morrison, Miranda, Lin, Chandu, Dziri, Kumar, Zick, Choi et~al.}]{RewardBench}
Nathan Lambert, Valentina Pyatkin, Jacob Morrison, LJ~Miranda, Bill~Yuchen Lin, Khyathi Chandu, Nouha Dziri, Sachin Kumar, Tom Zick, Yejin Choi, et~al. 2024.
\newblock Rewardbench: Evaluating reward models for language modeling.
\newblock \emph{arXiv preprint arXiv:2403.13787}.

\bibitem[{Li et~al.(2024)Li, Chiang, Frick, Dunlap, Wu, Zhu, Gonzalez, and Stoica}]{ArenaHard}
Tianle Li, Wei-Lin Chiang, Evan Frick, Lisa Dunlap, Tianhao Wu, Banghua Zhu, Joseph~E Gonzalez, and Ion Stoica. 2024.
\newblock From crowdsourced data to high-quality benchmarks: Arena-hard and benchbuilder pipeline.
\newblock \emph{arXiv preprint arXiv:2406.11939}.

\bibitem[{Li et~al.(2023)Li, Zhang, Dubois, Taori, Gulrajani, Guestrin, Liang, and Hashimoto}]{AlpacaEval}
Xuechen Li, Tianyi Zhang, Yann Dubois, Rohan Taori, Ishaan Gulrajani, Carlos Guestrin, Percy Liang, and Tatsunori~B. Hashimoto. 2023.
\newblock Alpacaeval: An automatic evaluator of instruction-following models.
\newblock \url{https://github.com/tatsu-lab/alpaca_eval}.

\bibitem[{Lin et~al.(2022)Lin, Hilton, and Evans}]{lin2022truthfulqa}
Stephanie Lin, Jacob Hilton, and Owain Evans. 2022.
\newblock {TruthfulQA}: Measuring how models mimic human falsehoods.
\newblock In \emph{Proceedings of the 60th Annual Meeting of the Association for Computational Linguistics (Volume 1: Long Papers)}, pages 3214--3252.

\bibitem[{Meng et~al.(2024)Meng, Xia, and Chen}]{SimPO}
Yu~Meng, Mengzhou Xia, and Danqi Chen. 2024.
\newblock Sim{PO}: Simple preference optimization with a reference-free reward.
\newblock In \emph{Advances in Neural Information Processing Systems}.

\bibitem[{Ouyang et~al.(2022)Ouyang, Wu, Jiang, Almeida, Wainwright, Mishkin, Zhang, Agarwal, Slama, Ray et~al.}]{InstructGPT}
Long Ouyang, Jeffrey Wu, Xu~Jiang, Diogo Almeida, Carroll Wainwright, Pamela Mishkin, Chong Zhang, Sandhini Agarwal, Katarina Slama, Alex Ray, et~al. 2022.
\newblock Training language models to follow instructions with human feedback.
\newblock \emph{Advances in Neural Information Processing Systems}.

\bibitem[{Pace et~al.(2024)Pace, Mallinson, Malmi, Krause, and Severyn}]{westofn}
Alizée Pace, Jonathan Mallinson, Eric Malmi, Sebastian Krause, and Aliaksei Severyn. 2024.
\newblock West-of-n: Synthetic preferences for self-improving reward models.
\newblock \emph{arXiv preprint arXiv:2401.12086}.

\bibitem[{Rafailov et~al.(2023)Rafailov, Sharma, Mitchell, Manning, Ermon, and Finn}]{DPO}
Rafael Rafailov, Archit Sharma, Eric Mitchell, Christopher~D Manning, Stefano Ermon, and Chelsea Finn. 2023.
\newblock Direct preference optimization: Your language model is secretly a reward model.
\newblock \emph{Advances in Neural Information Processing Systems}.

\bibitem[{Rosset et~al.(2024)Rosset, Cheng, Mitra, Santacroce, Awadallah, and Xie}]{DNO}
Corby Rosset, Ching-An Cheng, Arindam Mitra, Michael Santacroce, Ahmed Awadallah, and Tengyang Xie. 2024.
\newblock Direct nash optimization: Teaching language models to self-improve with general preferences.
\newblock \emph{arXiv preprint arXiv:2404.03715}.

\bibitem[{Schulman et~al.(2017)Schulman, Wolski, Dhariwal, Radford, and Klimov}]{PPO}
John Schulman, Filip Wolski, Prafulla Dhariwal, Alec Radford, and Oleg Klimov. 2017.
\newblock Proximal policy optimization algorithms.
\newblock \emph{arXiv preprint arXiv:1707.06347}.

\bibitem[{Singh et~al.(2024)Singh, Co-Reyes, Agarwal, Anand, Patil, Liu, Harrison, Lee, Xu, Parisi et~al.}]{ReSTEM}
Avi Singh, John~D Co-Reyes, Rishabh Agarwal, Ankesh Anand, Piyush Patil, Peter~J Liu, James Harrison, Jaehoon Lee, Kelvin Xu, Aaron Parisi, et~al. 2024.
\newblock Beyond human data: Scaling self-training for problem-solving with language models.
\newblock \emph{Transactions on Machine Learning Research}.

\bibitem[{Touvron et~al.(2023)Touvron, Martin, Stone, Albert, Almahairi, Babaei, Bashlykov, Batra, Bhargava, Bhosale et~al.}]{Llama2}
Hugo Touvron, Louis Martin, Kevin Stone, Peter Albert, Amjad Almahairi, Yasmine Babaei, Nikolay Bashlykov, Soumya Batra, Prajjwal Bhargava, Shruti Bhosale, et~al. 2023.
\newblock Llama 2: Open foundation and fine-tuned chat models.
\newblock \emph{arXiv preprint arXiv:2307.09288}.

\bibitem[{Wu et~al.(2024)Wu, Yuan, Golovneva, Xu, Tian, Jiao, Weston, and Sukhbaatar}]{Meta-rewarding}
Tianhao Wu, Weizhe Yuan, Olga Golovneva, Jing Xu, Yuandong Tian, Jiantao Jiao, Jason Weston, and Sainbayar Sukhbaatar. 2024.
\newblock Meta-rewarding language models: Self-improving alignment with llm-as-a-meta-judge.
\newblock \emph{arXiv preprint arXiv:2407.19594}.

\bibitem[{Wu et~al.(2025)Wu, Sun, Yuan, Ji, Yang, and Gu}]{SPPO}
Yue Wu, Zhiqing Sun, Huizhuo Yuan, Kaixuan Ji, Yiming Yang, and Quanquan Gu. 2025.
\newblock Self-play preference optimization for language model alignment.
\newblock In \emph{The Thirteenth International Conference on Learning Representations}.

\bibitem[{Xie et~al.(2025)Xie, Foster, Krishnamurthy, Rosset, Awadallah, and Rakhlin}]{XPO}
Tengyang Xie, Dylan~J Foster, Akshay Krishnamurthy, Corby Rosset, Ahmed~Hassan Awadallah, and Alexander Rakhlin. 2025.
\newblock Exploratory preference optimization: Provably sample-efficient exploration in {RLHF} with general function approximation.
\newblock In \emph{The Thirteenth International Conference on Learning Representations}.

\bibitem[{Xiong et~al.(2024)Xiong, Dong, Ye, Wang, Zhong, Ji, Jiang, and Zhang}]{xiong2024iterative}
Wei Xiong, Hanze Dong, Chenlu Ye, Ziqi Wang, Han Zhong, Heng Ji, Nan Jiang, and Tong Zhang. 2024.
\newblock Iterative preference learning from human feedback: Bridging theory and practice for rlhf under kl-constraint.
\newblock In \emph{International Conference on Machine Learning}.

\bibitem[{Xu et~al.(2023)Xu, Lee, Sukhbaatar, and Weston}]{PCO}
Jing Xu, Andrew Lee, Sainbayar Sukhbaatar, and Jason Weston. 2023.
\newblock Some things are more cringe than others: Preference optimization with the pairwise cringe loss.
\newblock \emph{arXiv preprint arXiv:2312.16682}.

\bibitem[{Yuan et~al.(2024)Yuan, Pang, Cho, Li, Sukhbaatar, Xu, and Weston}]{Self-rewarding}
Weizhe Yuan, Richard~Yuanzhe Pang, Kyunghyun Cho, Xian Li, Sainbayar Sukhbaatar, Jing Xu, and Jason~E Weston. 2024.
\newblock Self-rewarding language models.
\newblock In \emph{International Conference on Machine Learning}.

\bibitem[{Zellers et~al.(2019)Zellers, Holtzman, Bisk, Farhadi, and Choi}]{zellers2019hellaswag}
Rowan Zellers, Ari Holtzman, Yonatan Bisk, Ali Farhadi, and Yejin Choi. 2019.
\newblock {H}ella{S}wag: Can a machine really finish your sentence?
\newblock In \emph{Proceedings of the 57th Annual Meeting of the Association for Computational Linguistics}, pages 4791--4800.

\bibitem[{Zhang et~al.(2024{\natexlab{a}})Zhang, Zhoubian, Hu, Yue, Dong, and Tang}]{ReST-MCTS*}
Dan Zhang, Sining Zhoubian, Ziniu Hu, Yisong Yue, Yuxiao Dong, and Jie Tang. 2024{\natexlab{a}}.
\newblock Re{ST}-{MCTS}*: {LLM} self-training via process reward guided tree search.
\newblock In \emph{Advances in Neural Information Processing Systems}.

\bibitem[{Zhang et~al.(2024{\natexlab{b}})Zhang, Yu, Sharma, Yang, Wang, Hassan, and Wang}]{SELM}
Shenao Zhang, Donghan Yu, Hiteshi Sharma, Ziyi Yang, Shuohang Wang, Hany Hassan, and Zhaoran Wang. 2024{\natexlab{b}}.
\newblock Self-exploring language models: Active preference elicitation for online alignment.
\newblock \emph{arXiv preprint arXiv:2405.19332}.

\bibitem[{Zheng et~al.(2023)Zheng, Chiang, Sheng, Zhuang, Wu, Zhuang, Lin, Li, Li, Xing et~al.}]{LLM-as-a-Judge}
Lianmin Zheng, Wei-Lin Chiang, Ying Sheng, Siyuan Zhuang, Zhanghao Wu, Yonghao Zhuang, Zi~Lin, Zhuohan Li, Dacheng Li, Eric Xing, et~al. 2023.
\newblock Judging llm-as-a-judge with mt-bench and chatbot arena.
\newblock \emph{Advances in Neural Information Processing Systems}.

\bibitem[{Zheng et~al.(2024)Zheng, Shen, Hua, Lai, Dou, Zhou, Xi, Wang, Huang, Gui, Zhang, and Huang}]{zheng2024improving}
Rui Zheng, Wei Shen, Yuan Hua, Wenbin Lai, Shihan Dou, Yuhao Zhou, Zhiheng Xi, Xiao Wang, Haoran Huang, Tao Gui, Qi~Zhang, and Xuanjing Huang. 2024.
\newblock Improving generalization of alignment with human preferences through group invariant learning.
\newblock In \emph{The Twelfth International Conference on Learning Representations}.

\end{thebibliography}

\newpage
\appendix

\section{Baselines}
\label{Baselines}
We compare our approach against the following baseline methods.~\emph{Offline preference optimization} methods:
For this category, we implement DPO~\cite{DPO}, IPO ~\cite{IPO} and SimPO~\cite{SimPO}. Preference pairs are derived from multiple responses generated by the base policy model, with scores provided by the base reward model.
\emph{Iterative preference optimization} methods:
For this category, we implement iterative DPO \cite{PCO}, SPPO~\cite{SPPO} and Meta-Rewarding~\cite{Meta-rewarding}. Since these methods do not update the reward model, we use all three portions of the dataset for policy model training and run three iterations for iterative methods, i.e., SPPO and Meta-Rewarding. To ensure a fair comparison, the sampling settings used in these experiments match those applied in Mutual-Taught.

\section{Training Details}
\label{Experiments Details}
In our experiments, we use the Alignment Handbook framework\footnote{Alignment Handbook at \url{https://github.com/huggingface/alignment-handbook}} for policy model updates and the RLHF-Reward-Modeling\footnote{RLHF-Reward-Modeling at \url{https://github.com/RLHFlow/RLHF-Reward-Modeling}} framework for reward model updates.

\paragraph{Mutual-Taught} We conduct Mutual-Taught between the policy and reward models for two iterations. In each iteration, both models are trained for one epoch using a cosine learning rate schedule with a warmup ratio of 0.1. All experiments are conducted on 8 NVIDIA A100 GPUs.
We follow SimPO \cite{SimPO} to set the policy sampling and training parameters. Specifically, for policy sampling: the temperature is set to 0.8, \(M=5\), and top-p to 0.95. 
For each policy model iteration, we initialize the model from the previous round and generate responses using the current policy. Preference data is then derived using the reward model at the current iteration. The policy model is optimized via DPO with a beta of 0.01, a batch size of 128, a maximum sequence length of 2,048 tokens, and a learning rate of \(7 \times 10^{-7}\).
A checkpoint is saved every 50 steps for subsequent model selection. For model selection, a fixed evaluation set is constructed prior to the start of the iterations by randomly sampling 2,000 prompts from the UltraFeedback dataset. Among the saved checkpoints, the one with the highest win-rate relative to the initial policy of the current iteration is selected to construct the pseudo-labels. The iteration is terminated if the highest win-rate $w_t^k$ is less than 60\%. 
For data filtering, the margin threshold is set based on the standard deviation of the reward model scores in the current iteration. 

To mitigate the risk of overfitting on the same prompts across iterations, \emph{each reward model iteration starts from the base reward model}. The reward model is trained on preference pairs consisting of chosen and rejected responses sampled from the current and preceding policy models. We use a batch size of 512, a maximum sequence length of 2,048, and a learning rate of \(2 \times 10^{-6}\).

\paragraph{Baselines} In \textit{offline preference optimization} methods, we maintain the same sampling and training parameters as Mutual-Taught. For \textit{iterative preference optimization} methods, in iterative DPO, we observed performance degradation in the final iteration with a large learning rate, so we lowered it to \(5 \times 10^{-7}\). For SPPO, we use the default training parameters provided by the method. For Meta-Rewarding, we first build Evaluation Fine-Tuning (EFT) data from the Open Assistant \cite{OpenAssistant} dataset to boost the initial judgment ability of the model before self-training iterations. During the construction of EFT data, we prompt GPT-4o to generate judgments with high quality instead of the SFT baseline in \citeauthor{Self-rewarding} \citeyearpar{Self-rewarding}. During self-training iterations, we use prompts from the UltraFeedback dataset instead of those generated by Llama2-70B-Chat to align with Mutual-Taught.

\paragraph{Length control} To prevent length explosion, we implement a length-control mechanism for selecting preference data. For each prompt, we first select responses with above-average reward scores, and then choose the shortest one as the chosen response. The response with the lowest score is selected as the rejected one. This length control mechanism is applied to all experiments except for Meta-Rewarding, where we use the length control mechanism proposed by the original method.

\section{Algorithmic Overview}
Algorithm~\ref{alg} outlines the complete Mutual-Taught procedure. In classical EM, both the variational approximation of the latent variable and the model parameters are iteratively refined. Analogously, we treat \(\pi^*\) as the latent variable and the policy \(\pi_t\) as an evolving surrogate. 
By refining the policy in the E-step and adjusting the reward model in the M-step, both models progressively align with the latent optimal distribution \(\pi^*\).

\begin{algorithm}[htbp]
\caption{Mutual-Taught}
\label{alg}
\begin{algorithmic}[1]
\STATE \textbf{Input:} Initial policy \(\pi_0\), initial reward model \(r_0\), dataset \(\mathcal{D}\), fixed validation set \(D_\text{MS}\), number of iterations \(T\).
\STATE Partition \(\mathcal{D}\) into subsets \(\mathcal{D}_1, \ldots, \mathcal{D}_T, \mathcal{D}_R\), where \(\mathcal{D}_1\) to \(\mathcal{D}_T\) are used for policy model updates, and \(\mathcal{D}_R\) is utilized for reward model updates. Additionally, \(\mathcal{D}_\text{MS}\) is designated for model selection.
\FOR{each iteration \(t = 1, \ldots, T\)}
    \STATE \textbf{E-step:} Obtain policy checkpoints \(\{\pi'_t\}\) by sampling responses from \(\pi_{t-1}\) for \(x \sim \mathcal{D}_t\), evaluating them with \(r_{t-1}\), and updating \(\pi_{t-1}\) according to Eq.~(\ref{eq:e-step}).
    \STATE \textbf{Model selection:} Select the best policy \(\pi_t\) via Eq.~(\ref{eq:model_sel}).
    \STATE \textbf{Pseudo-pair construction:} For each prompt \(x \sim \mathcal{D}_R\), construct the pseudo-pair \((y_t, y_{t-1})\) by generating \(y_t \sim \pi_t(x)\) as the preferred response and \(y_{t-1} \sim \pi_{t-1}(x)\) as the dispreferred response.
    \STATE \textbf{Data filtering:} Discard the pseudo-pair if it does not satisfy the margin threshold $\epsilon_t$.
    \STATE \textbf{M-step:} Update \(r_{t-1}\) using the filtered pseudo-pairs according to Eq.~(\ref{eq:m-step}).

\ENDFOR
\STATE \textbf{Output:} Policy \(\pi_T\) and reward model \(r_T\).
\end{algorithmic}
\end{algorithm}

\section{Theoretical Convergence Analysis}
\label{appendix:theoretical_convergence}

The Mutual-Taught algorithm draws theoretical inspiration from the classical Expectation-Maximization (EM) framework while introducing novel components. Under standard regularity conditions, we establish its convergence properties through the following formal analysis.

\subsection{Objective Formulation}
Let the expected reward under the latent optimal distribution be defined as:
\[
R(\pi^*, r) = \mathbb{E}_{x \sim \mathcal{D},\, y \sim \pi^*(\cdot \mid x)} \bigl[r(y; x)\bigr],
\]
where $\pi^*$ represents the ground-truth distribution of optimal responses. Our convergence analysis focuses on the sequence $\{(\pi_t, r_t)\}_{t=1}^T$ generated by alternating optimization steps.

\subsection{Convergence Theorem}
\begin{theorem}[Monotonic Improvement]
Under the assumptions that:
\begin{enumerate}
    \item Exact optimization in E-step and M-step.
    \item Unbiased estimation in pseudo-labeling: $\mathbb{E}[\hat{\pi}(y|x)] = \pi^*(y|x)$.
\end{enumerate}
The Mutual-Taught sequence satisfies:
\[
R(\pi_t, r_t) \geq R(\pi_{t-1}, r_{t-1})\quad \forall t \geq 0,
\]
with equality holding if and only if $(\pi_t, r_t) = (\pi_{t-1}, r_{t-1})$. Thus, the algorithm converges to a stationary point of $R(\pi, r)$, ensuring asymptotic convergence to a solution where no further improvement is possible.
\end{theorem}

\subsection{Proof Sketch}
The convergence follows from alternating maximization principles, with two key enhancements:

\begin{enumerate}

\item \textbf{E-step:~Progressive policy improvement via model selection}
\begin{itemize}
\item The policy update maximizes the auxiliary lower bound:
\[
R(\pi, r_{t-1}) \geq \mathbb{E}\left[\log \pi(y|x)r_{t-1}(y;x)\right].
\]
\item Model selection ensures non-degeneracy: By monitoring validation set performance, we ensure that the new policy update satisfies:
\[
R(\pi_t, r_{t-1}) \geq R(\pi_{t-1}, r_{t-1}).
\]
\item Selection mechanism prevents performance regression by discarding suboptimal policy updates.
\end{itemize}

\item \textbf{M-step:~Progressive reward model enhancement with data filtering}
\begin{itemize}
\item The reward model is updated by maximizing the pairwise preference likelihood as follows:
\[
\max_r \mathbb{E}_{(y_w, y_l)\sim \hat{\pi}} \log\sigma(r(y_w; x)-r(y_l;x)).
\]
\item Margin-based filtering enforces quality control: since low-quality pairs are discarded, we ensure that the new reward model satisfies:
\[
R(\pi_{t}, r_{t}) \geq R(\pi_{t}, r_{t-1}).
\]
\item This data filtering strategy ensures $\text{Cov}(\hat{\pi})\!\to\!\text{Cov}(\pi^*)$, thereby reducing approximation error and enhancing the accuracy of the reward model.
\end{itemize}
\end{enumerate}

The joint effect of these steps can be captured by the chained inequalities:
\[
R(\pi_t, r_t) \overset{\text{M-step}}{\geq} R(\pi_t, r_{t-1}) \overset{\text{E-step}}{\geq} R(\pi_{t-1}, r_{t-1}).
\]
The two-stage stabilization strategy with model selection and data filtering essentially converts the original non-convex problem into a sequence of convex subproblems with progressively tightened constraints. This approach distinguishes Mutual-Taught from vanilla EM implementations, enabling more reliable convergence while preserving the original framework's theoretical benefits.

\begin{figure}[htbp]
    \centering
    \includegraphics[width=1\linewidth]{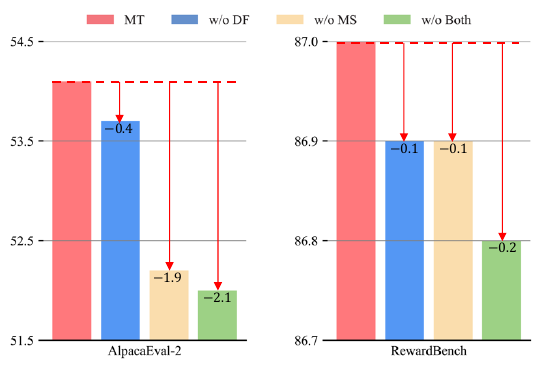}
    \caption{Ablation study on the two-stage strategy. For brevity, Mutual-Taught, model selection and data filtering are abbreviated as MT, MS and DF, respectively.}
    \label{fig:ablation-study}
\end{figure}

\section{Ablation Studies of Two-Stage Stabilization}
\label{appendix:ablation_twostage}

To demonstrate the effectiveness of the proposed two-stage stabilization strategy, we conduct an ablation study. As shown in Figure \ref{fig:ablation-study}, we draw two key observations:
\begin{itemize}
    \item Both model selection and data filtering individually improve performance over the baseline without the two-stage strategy (i.e., ``w/o Both''), indicating that each component effectively enhances pseudo-label quality.
    \item While model selection and data filtering confer similar benefits to the reward model, model selection provides a greater advantage for policy model optimization. This is because the policy selected according to Eq.~(\ref{eq:e-step}) not only yields more reliable pseudo-labels for the M-step but also serves as a better initialization for the next policy update.
\end{itemize}


\section{Pseudo-Label Filtering Methods}
\label{appendix:different_filter_methods}

As demonstrated in Appendix~\ref{appendix:ablation_twostage}, the performance of Mutual-Taught critically depends on the quality of its pseudo-labels. To reduce noise in the generated preference pairs, we systematically analyze three curation strategies:

\begin{itemize}
    \item \emph{Low-Quality Data Filtering} (LQF): Eliminate pseudo-pairs where the preferred response $y_t$ scores \emph{lower} than the dispreferred response $y_{t-1}$ by a margin: $\Delta r(x) < -\epsilon_t$.

    \item \emph{High-Quality Data Selection} (HQS): Retain only pseudo-pairs in which the preferred response $y_t$ scores \emph{higher} than the dispreferred response $y_{t-1}$ by a margin: $\Delta r(x) \geq \epsilon_t$.
    
    \item \emph{Direct Self-Training} (DST): Directly compare reward model scores of the pre- and post-update policy responses, designating the higher-scoring response as preferred.
\end{itemize}

\begin{figure}[htbp]
    \centering
    \includegraphics[width=1\linewidth]{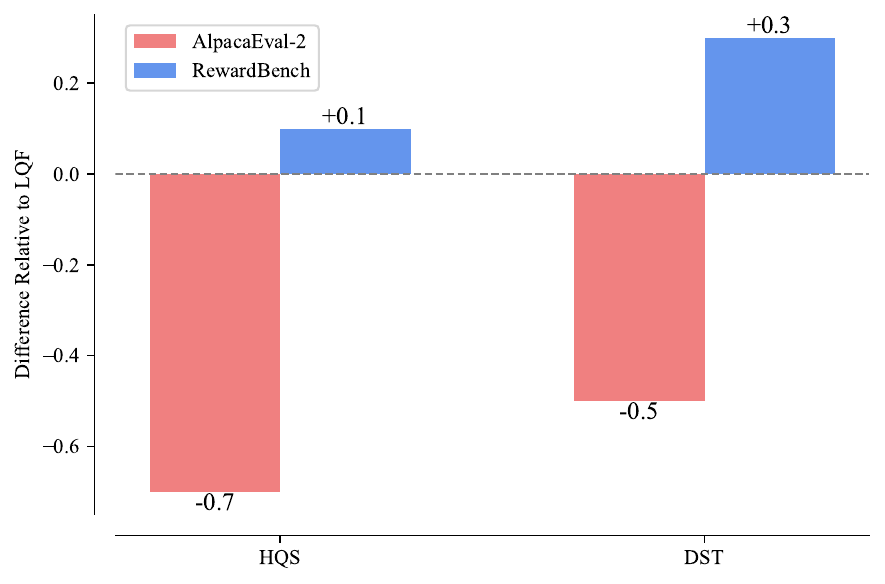}
    \caption{Comparison of different data filtering methods. The vertical axis displays the performance differences of High-Quality Data Selection (HQS) and Direct Self-Training (DST) relative to Low-Quality Data Filtering (LQF) on two benchmarks.}
    \label{fig:diff_filtering_method}
\end{figure}

Figure~\ref{fig:diff_filtering_method} shows that while LQF (our adopted approach in the final method) delivers superior performance on AlpacaEval-2, HQS and DST slightly outperform it on RewardBench. By analyzing their underlying mechanisms, we observe:

\begin{itemize}
    \item \emph{Both HQS and DST are essentially self-training approaches.} While self-training can alleviate catastrophic forgetting (Section~\ref{Impact of Reward Model Training Data Type}), it effectively enhances the existing capabilities of the reward model. However, for samples where the reward model fails to correctly recognize due to policy distribution shift, self-training alone may not provide the necessary calibration signals. In contrast, LQF filters out only the high-confidence low-quality samples, retaining data containing calibration information based on the comparison between pre- and post-update policies. This enables the reward model to provide more accurate feedback for subsequent policy improvements. 
    
    \item \emph{HQS can be viewed as a special case of DST,} where only responses that are strictly better under the updated policy are retained. In contrast, DST uses \emph{all} pseudo-labeled data, which leverages the reward model’s strong initial capacity. However, when the reward model’s initial capability is weaker, relying solely on self-training may lead to suboptimal behavior. In our case, since FsfairX-Llama3-RM-v0.1 has a strong initialization, DST achieves better performance on the reward model.
\end{itemize}

\section{Evaluation on Additional Benchmarks}
To further assess the effectiveness of Mutual-Taught across diverse downstream tasks and evaluation metrics, we conducted additional experiments on four benchmarks from the HuggingFace Open LLM Leaderboard ~\citep{open-llm-leaderboard}: GSM8K~\citep{cobbe2021gsm8k}, MMLU~\citep{hendrycks2021measuring}, HellaSwag~\citep{zellers2019hellaswag}, and TruthfulQA~\citep{lin2022truthfulqa}. The results are summarized in Table~\ref{tab:extra-benchmarks}. 

\begin{table}[h]
\centering\small
\resizebox{\columnwidth}{!}{%
\begin{tabular}{lccccc}
\toprule
\textbf{Model} & \textbf{GSM8K} & \textbf{MMLU} & \textbf{HellaSwag} & \textbf{TruthfulQA} & \textbf{Avg.} \\
\midrule
Base               & 75.21 & 65.71 & 78.48 & 51.64 & 67.76 \\
+ IterDPO         & 69.71 & 65.19 & 80.83 & 52.91 & 67.16 \\
+ \textsc{MT}      & 70.67 & 64.13 & \textbf{81.37} & \textbf{55.21} & \textbf{67.85} \\
\bottomrule
\end{tabular}
}
\caption{Accuracy (\%) on additional benchmarks from the HuggingFace Open LLM Leaderboard. Base refers to Llama-3-8B-Instruct; + IterDPO and + MT indicate models fine-tuned with Iterative DPO and Mutual-Taught, respectively.}
\label{tab:extra-benchmarks}
\end{table}

As shown in Table~\ref{tab:extra-benchmarks}, all preference optimization methods show performance drops on MMLU and GSM8K—likely due to the UltraFeedback dataset's emphasis on alignment over general knowledge and mathematics. In contrast, there is a consistent improvement on HellaSwag and TruthfulQA. These results suggest that the UltraFeedback dataset is more aligned with tasks requiring commonsense reasoning and truthfulness, and that Mutual-Taught is particularly beneficial in these areas.

\section{Threshold Selection for $\tau$ in the E-Step}

During early training, the policy model (PM) typically improves markedly after each E-step, reflected by validation win rates well above 50\%. In this regime, the updated response $y_t$ almost always surpasses its predecessor $y_{t-1}$. Meanwhile, the data-filtering procedure in the M-step discards unreliable preference pairs, keeping the reward model (RM) aligned with the evolving PM distribution.
As optimization advances, incremental gains taper off and the win rate converges toward 50\%. Distinguishing successive policies then becomes difficult, and marginally noisy pairs may impair the RM. To prevent over-optimization while preserving meaningful updates, we introduce a win-rate threshold~$\tau$: \textit{Above 50\% to prevent over optimization, while not excessively high to ensure continued meaningful optimization.}

To further evaluate $\tau$'s effectiveness in triggering timely early stopping during performance degradation, we extended our experiment by dividing the original set of 40,000 prompts (previously used in two iterations) into four subsets and conducting four iterations of PM training. The results are summarized in Table~\ref{tab:tau}. 

\begin{table}[h]
    \centering
    \resizebox{0.9\linewidth}{!}{%
    \begin{tabular}{cccc}
        \toprule
        \textbf{Iter.} & \textbf{MS Win (\%)} & \textbf{ES} & \textbf{AlpacaEval-2 LC (\%)} \\
        \midrule
        1 & 63.5 & No  & 34.7 \\
        2 & 67.3 & No  & 41.0 \\
        3 & 65.1 & No  & 44.9 \\
        4 & 57.7 & \checkmark & 40.3 \\
        \bottomrule
    \end{tabular}
    }
    \caption{Impact of early-stop threshold $\tau$ across iterations. "MS Win" denotes the win rate (\%) of the selected model in model selection (MS), "ES" indicates whether early stopping (ES) was applied during the iteration.}

    \label{tab:tau}
\end{table}

As shown in the Table \ref{tab:tau}, the PM exhibits performance degradation in the fourth iteration. By setting $\tau = 60\%$, based on the win rate (63.5\%) from the first PM iteration, early stopping was successfully triggered in the fourth iteration. Consequently, the model from the third EM iteration was selected as the final model. This demonstrates how $\tau$ effectively ensures early stopping at the appropriate point when performance degradation is detected.

\end{document}